\documentclass[letterpaper, 10 pt, journal, twoside]{ieeetran}

\IEEEoverridecommandlockouts                              
\usepackage{listings}
\usepackage[dvipsnames]{xcolor}
\usepackage{amsmath}
\usepackage{amssymb}
\usepackage{breqn}
\usepackage{algorithm}
\usepackage[noend]{algpseudocode}
\usepackage{gensymb}
\usepackage{graphicx}
\usepackage{amsmath}
\usepackage{color}
\usepackage[mathscr]{euscript}
\usepackage{float}

\usepackage{multicol}

\usepackage[mathscr]{euscript}

\usepackage{capt-of,etoolbox}

\newcommand\blfootnote[1]{%
  \begingroup
    \renewcommand\thefootnote{\fnsymbol{footnote}}%
    \renewcommand\thempfootnote{\fnsymbol{mpfootnote}}%
    \footnotetext[0]{#1}%
  \endgroup
}

\newcommand{\andy}[1]{#1}

\title{\LARGE \bf
Complex In-Hand Manipulation via Compliance-Enabled Finger Gaiting and Multi-Modal Planning}

\author{Andrew S. Morgan$^{1*}$, Kaiyu Hang$^{2*}$, Bowen Wen$^3$, Kostas Bekris$^3$, and Aaron M. Dollar$^1$
\vspace{-0.35in}

}

\def\NoNumber#1{{\def\alglinenumber##1{}\State #1}\addtocounter{ALG@line}{-1}}
\AtBeginDocument{\setlength\abovedisplayskip{4pt}}
\AtBeginDocument{\setlength\belowdisplayskip{4pt}}

\begin{document}

\twocolumn[{
\renewcommand\twocolumn[1][]{#1}%
\maketitle

\begin{center} 
    \centering
     \includegraphics[width = 0.87\textwidth]{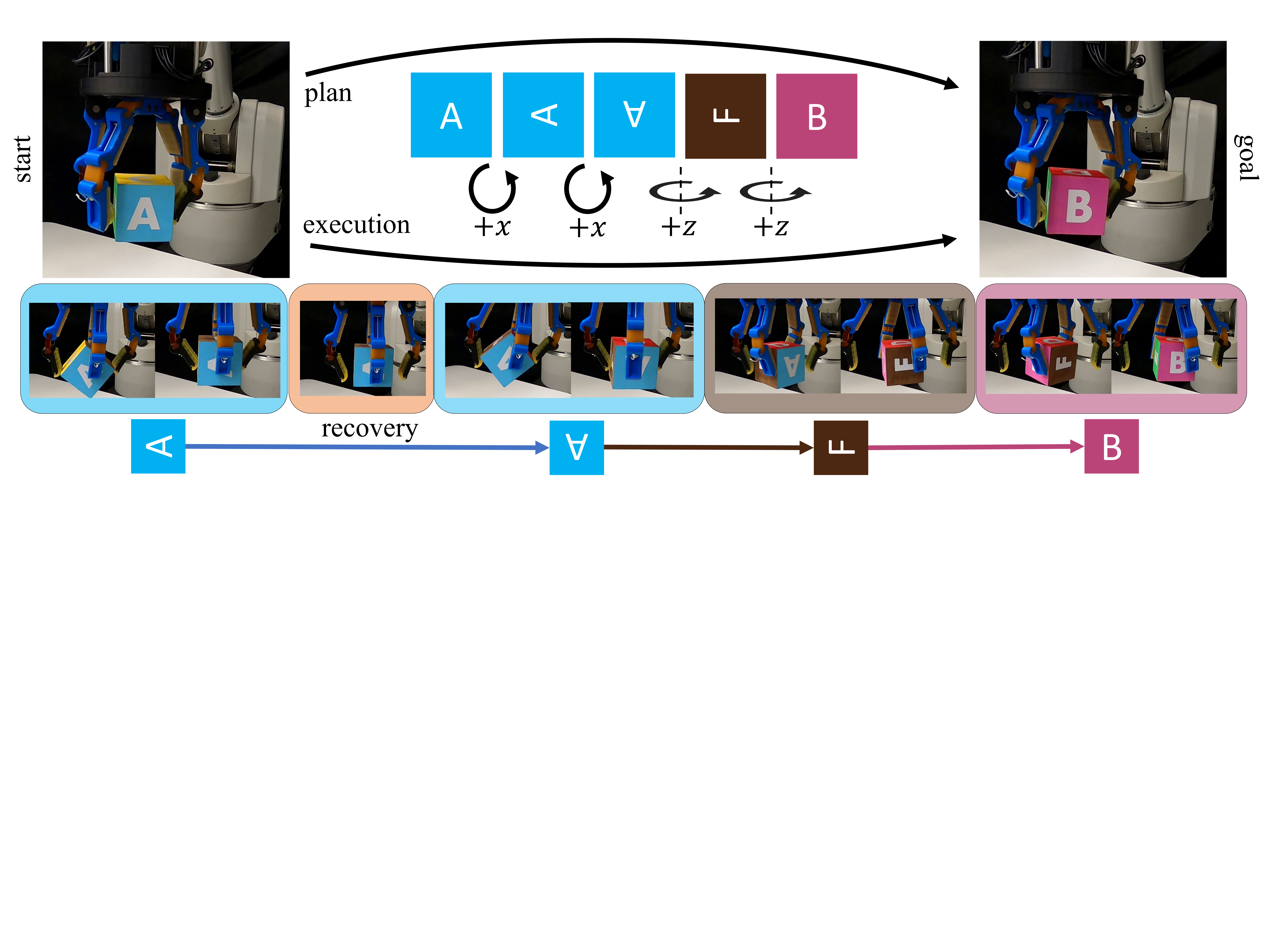}
      \vspace{-0.15in}
    \captionof{figure}{We explore the development of a complete $SO(3)$ planner for within-hand manipulation using finger gaits, by controlling two orthogonal extrinsic rotation axes. Given a start configuration (cube face A), the proposed planner finds an action sequence along the two controlled dimensions so as to reach the desired goal configuration (cube face B). During manipulation, the pose of the object is tracked via a low-latency, 6D pose object tracker, providing feedback for online replanning and disturbance compensation via a recovery phase that uses translation control.}
    \vspace{-.05in}
\end{center}%
}]

\blfootnote{Manuscript received: September 9, 2021; Revised: December 17, 2021; Accepted: January 6, 2022.}
\blfootnote{This paper was recommended for publication by Editor Hong Liu upon evaluation of the Associate Editor and Reviewers’ comments.}
\blfootnote{This work was supported by the U.S. National Science Foundation grants IIS-1752134, IIS-1734190, \& IIS-1900681 to Yale University and NRI-1734492 to Rutgers University. * Authors contributed equally.}
\blfootnote{$^1$Andrew Morgan and Aaron Dollar are with the Department of Mechanical Engineering \& Materials Science, Yale University, USA. (email: {\tt\footnotesize {Andrew.Morgan, Aaron.Dollar\}}@yale.edu)}}
\blfootnote{$^2$Kaiyu Hang is with the Department of Computer Science, Rice University, USA. (email: {\tt\footnotesize kaiyu.hang@rice.edu)}}
\blfootnote{$^3$Bowen Wen and Kostas Bekris are with the Department of Computer Science, Rutgers University, USA. (email: {\tt\footnotesize \{bw344, kostas.bekris\}@cs.rutgers.edu)}}
\blfootnote{Digital Object Identifier (DOI): see top of this page.}
 
\vspace{-.15in}
\begin{abstract}

\andy{Constraining contacts to remain fixed on an object during manipulation limits the potential workspace size, as motion is subject to the hand's kinematic topology. Finger gaiting is one way to alleviate such restraints. It allows contacts to be freely broken and remade so as to operate on different manipulation manifolds. This capability, however, has traditionally been difficult or impossible to practically realize. A finger gaiting system must simultaneously plan for and control forces on the object while maintaining stability during contact switching. This work alleviates the traditional requirement
by taking advantage of system compliance, allowing the hand to more easily switch contacts while maintaining a stable grasp. Our method achieves complete $SO(3)$ finger gaiting control of grasped objects against gravity by developing a manipulation planner that operates via orthogonal safe modes of a compliant, underactuated hand absent of tactile sensors or joint encoders. 
During manipulation, a low-latency 6D pose object tracker provides feedback via vision, allowing the planner to update its plan online so as to adaptively recover from trajectory deviations. 
The efficacy of this method is showcased by manipulating both convex and non-convex objects on a real robot. Its robustness is evaluated via perturbation rejection and long trajectory goals. To the best of the authors' knowledge, this is the first work that has autonomously achieved full $SO(3)$ control of objects within-hand via finger gaiting and without a support surface, elucidating a valuable step towards realizing true robot in-hand manipulation capabilities.} 



\end{abstract}

\vspace{-.15in}

\vspace{-0.15in}
\section{INTRODUCTION}

\IEEEPARstart{W}{ithin-Hand Manipulation} can be characterized as the ability to reorient or reposition an object with respect to the hand frame \cite{bicchi2000hands}. The quest for this capability has been studied in the robot manipulation community for decades--from planning and control with rigid hands \cite{kerr1986analysis} to efforts with soft, compliant, or underactuated hands \cite{sintov2019learning}. Notably, the majority of these works constrain contacts to remain fixed or rolling during manipulation. This constraint limits the object's workspace as the actuators can only operate on a single hand-object configuration manifold. Alternatively, finger gaiting, i.e., the process of repositioning contacts on the object during manipulation, can help alleviate such limitations and extend the object's available workspace. 

Finger gaiting is an inherently difficult task for a robot. Given a robot hand, the individual serial link fingers must work in proper unison without collision; maintaining stability while making and breaking contact with the object \cite{han1998dextrous}. The computational complexity of this problem has traditionally been very expensive--requiring the system to calculate and modulate forces and planned joint trajectories online during manipulation. We, conversely, are able to alleviate many of these complexities by leveraging compliance, i.e. safe modal transitions, in our system. \andy{ The capabilities we present extend beyond what has been illustrated previously in the literature, some purely in simulation \cite{sundaralingam2018geometric, khandate2021feasibility} and some demonstrated on a real robot, but with support surfaces \cite{andrychowicz2020learning, bhatt2021surprisingly, Abondance2020} . }



In this letter, we build off the observation that by leveraging the passive adaptive properties of an underactuated hand, we can convert traditional position, force, and stability control problems to a unified motion-only control paradigm, based on the compliant properties of the hand. Concretely, finger gaiting requires contacts to be constantly in motion. The presence and thus activation of certain contacts constrains what forces the hand can impart onto the object. This phenomena creates the abstraction of modes, which can be conceptualized as different families of motion manifolds that are subject to the system's current set of constraints--typically in the form of gaining or losing contacts. Multi-modal planning within and between these constraints thus becomes a major focus of this work. 

We in this work can constrain our planning approach according to the nature of the task. Formally, any orientation in $SO(3)$ can be achieved via a three action trajectory comprised of two orthogonal rotations. From this formulation, we design two core modal actions for the hand and a multi-modal planner that runs online and is constantly updated via a vision-based, low-latency 6D pose object tracker \cite{wen2020se, wen2020robust}. This method continually calculates a trajectory from start to goal and when undesired scenarios arise, such as contact slip, the planner updates and suggests new actions accordingly. We incorporate this approach into an open-source and underactuated hand with four fingers \cite{ma2014underactuated}. In the end, we showcase the efficacy of our system through various experiments: tracking the planned and executed trajectory of an object, evaluating the recovery potential given undesired perturbations, and finally, the ability to extend to novel object geometries. 

\andy{The contributions in this letter are threefold. First, we develop a \textit{complete} and fast planning solution for in-hand reorientation using two extrinsic rotation axes. Secondly, we describe the utility of compliance for switching between modes and how it ``inflates" the contact switching region. And, finally, we present a simple yet effective robot system capable of complex finger gaiting capabilities, underscoring continued discussion in the community on the utility of compliance for in-hand manipulation tasks. }
\vspace{-0.1in}

\section{RELATED WORK}

\vspace{-0.03in}
\subsubsection{Modeling Manipulation}
Modeling robot manipulation has historically been arduous as the dynamics associated with contact are difficult to predict in novel scenarios. From an in-hand manipulation perspective, various levels of modeling have been investigated -- from contact models \cite{TrinkleContact} and fingerpad curvature models \cite{Montana1992}, to hand kinematic models and whole hand-object system models \cite{Murray1994a}. While these approaches have elucidated many powerful hand-object relationships, inaccurate parameterizations can often lead to task failure \cite{hang2021manipulation}. To help alleviate such uncertainty, several end effectors leverage designs that are soft or underactuated \cite{Dollar2010a}. This property provides passive reconfigurability to the mechanism, which can absorb much of the ``slack" traditionally required to be fully accounted for in system modeling and can further reduce grasp planning times \cite{hang2019pre, morgan2021vision}. In this work, we leverage such mechanisms to unify our planning approach--creating the notion of safe regions for contact switching. 

\subsubsection{Extending In-hand Manipulation Capabilities}
Models for in-hand manipulation are typically constrained to fixed or rolling contact scenarios. While this advantageously simplifies assumptions for control, it also limits the object's available workspace to be dependent on the kinematic topology of the hand. Without relying on external contact, e.g, \cite{chavan-dafle2019ijrr}, \andy{ or task-specific hands with roller-based fingers, e.g \cite{yuan2020roller},} two general approaches can help alleviate such constraint: sliding manipulation and finger gaiting. Leveraging the former is very difficult, as detecting and controlling its nonlinear conditions requires various levels of advanced sensing \cite{brock1988enhancing, su2015force}. The latter, alternatively, has been largely difficult due to computational considerations, but has been made more successful in recent years \cite{Birchereabd2666, khandate2021feasibility}. 
\andy{
The work in \cite{hang2016hierarchical} used finger gaits and tactile sensors to maintain grasp stability. \cite{andrychowicz2020learning} used a 24-DOF hand with a motion capture system, in addition to ``over 100 years" of simulated data to perform impressive and fast cube manipulation in the palm. \cite{bhatt2021surprisingly} leveraged the capabilities of a soft hand with 16 degrees of actuation for similar types of manipulation. While as impressive as these aforementioned works are, we are interested in extending beyond these capabilities to work against gravity, i.e. maintaining object stability without a support surface, and while utilizing a simple hand with very little onboard sensing.}
  \begin{figure*}[t]
      \centering
      \includegraphics[width = 0.98\textwidth]{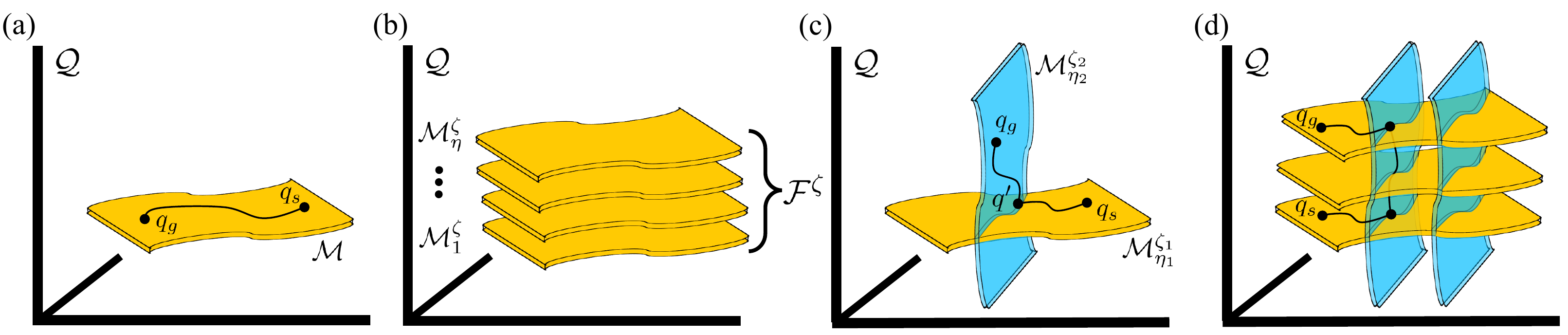}
       \vspace{-0.1in}
      \caption{Multi-modal planning problems can be conceptualized as operating in different configuration manifolds. (a) Given a single manifold $\mathcal{M}$, the planner must find a path along the constrained layer. (b) A single mode, or foliation, can have multiple manipulation manifolds depending on the start configuration of the hand-object system. (c) Switching modes is possible when the system finds a configuration, $q'$, that lies on both manifolds. (d) Finding a path from start configuration, $q_s$, to goal configuration, $q_g$, can require multiple jumps between modes. Note that although represented in this figure as folds, we assume modes to be differentiable but not necessarily euclidean. }
      \label{fig:mmPlanning}
      \vspace{-.2in}
   \end{figure*}
   
\subsubsection{Multi-Modal Planning for Manipulation}
Problems in robot manipulation are frequently multi-modal, i.e. the seemingly continuous problems have an underlying discrete structure guided by constraints \cite{hauser2010multiOG}. These constraints are typically imposed by the nature of contact, and by discretizing planning in terms of these manifolds, the planner search space is confined according to physical constraints \cite{zucker2013chomp}. Numerous sampling-based multi-modal planners have been described in the literature, 
which are able to generalize well, particularly in high-DOF scenarios. Though, sampling-based methods without informed exploration are vastly inefficient and suffer from time-complexities associated with over-exploration. Recent work has attempted to address this issue by using informed ``leads" to guide exploration \cite{kingston2020informing}. In our work, we build off these observations and constrain our planner's search according to physical properties of the $SO(3)$ rotation group. That is, our developed planner constrains the number and nature of mode switching to reduce planning time for continual updates during online execution.     




\vspace{-.1in}

\section{PRELIMINARIES} \label{sec:prelims}
In this section, we introduce the preliminaries associated with multi-modal motion planning. We first discuss constrained manifolds and then develop notation and terminology for modal switching leveraging compliance.  

\vspace{-.1in}
\subsection{Constrained Within-Mode Planning} \andy{Consider a robot with configuration space, $\mathcal{Q} \subset \mathbb{R}^N$, where $N$ is the number of joints and where $\mathcal{Q}$ can completely define the state of the robot.} Now, consider an object with configuration space, $\mathcal{O} \subset SE(3)$, which is located on a support surface, e.g. table top. For simplicity, let's disregard other potential collisions in the environment. In free space, the current robot configuration, $q \in \mathcal{Q}$, is unconstrained and thus able to freely move. Though, when $q$ causes links of the robot to come in contact with the object in its current configuration, $o \in \mathcal{O}$, a constraint is imposed on the system. The robot cannot push the object through the support surface. If, for instance, the goal of the robot is to slide the object along its support surface, the robot's motion is thus constrained to a contact configuration manifold, $\mathcal{M}$ (Fig. \ref{fig:mmPlanning}(a)). 

Assume there are several manifolds, indexed by $\eta$. The available motion manifold of the robot, $\mathcal{M}_\eta$, is subject to its associated constraint function $F^\eta(q,o):\mathbb{R}^N \rightarrow \mathbb{R}^{k_\eta}$ where $(1 \le k_\eta < N)$. Modes are thus defined as a family, or foliation, $\mathcal{F}^\zeta$
, of constrained manifolds that share the same, or similar, constraint functions 
(Fig. \ref{fig:mmPlanning}(b)). Inside each $\mathcal{F}^\zeta$, we assume all $\mathcal{M}_\eta$ to be smooth and differentiable. Therefore, planning in the $\mathcal{M}_\eta \in \mathbb{R}^{(N-k_\eta)}$ manifold of $\mathbb{R}^N$ is subject to, 
\begin{equation}\label{eq:manifold}
\mathcal{M}_\eta = \{q \in \mathcal{Q} |F^\eta(q,o) = \textbf{0} \}
\end{equation}

\vspace{-.15in}
\subsection{Safe Mode Transitions}
Multi-modal planning requires a robot to transition between at least two constraint manifolds, which likely lie in different foliations, in order to reach the desired goal. Consider a robot starting in configuration $q_{s}$, located in manifold $\mathcal{M}^{\zeta_1}_{\eta_1}$. Let's now assume the goal configuration $q_g$ is located in a different mode and thus along a different manifold, i.e. $\mathcal{M}^{\zeta_2}_{\eta_2}$ (Fig. \ref{fig:mmPlanning}(c)). In order for a robot to transition to $q_g$, a planner must find a candidate submanifold, $\mathcal{M}_{\eta_1 \cap \eta_2}$, that enables a transition. Inside of this submanifold lies candidate configuration, $q'$, for transitioning. We can define $\mathcal{M}_{\eta_1 \cap \eta_2}$ according to, 
\begin{equation}\label{eq:modes}
\mathcal{M}_{\eta_1 \cap \eta_2} = \{q \in \mathcal{Q} |F^{\eta_1}(q,o) = \textbf{0}, F^{\eta_2}(q,o) =\textbf{0} \}
\end{equation}
Note that it is possible for $\mathcal{M}_{\eta_1 \cap \eta_2}$ to be empty when there exists no $q'$ to transition $\mathcal{M}^{\zeta_1}_{\eta_1}$ to $\mathcal{M}^{\zeta_2}_{\eta_2}$.

In many robot scenarios, if a transition is possible from $\mathcal{M}^{\zeta_1}_{\eta_1}$ to $\mathcal{M}^{\zeta_2}_{\eta_2}$, the transfer region typically has a volume near zero, i.e. $\Omega(\mathcal{M}_{\eta_1 \cap \eta_2}) \approx 0$ where $\Omega$ is the convex hull function. There are situations, however, where we can relax this constraint to form \textit{safe mode transitions}. Formally, the manifold's constraint function, $F^\eta(q,o)$, is dependent on the robot's current configuration, $q$ and object configuration, $o$. Though due to near-unmodelable DOFs in soft, compliant, or underactuated robots or objects, the controlled configuration of the robot may still hold for $\mathbb{R}^N \rightarrow \mathbb{R}^{k_\eta}$, but the true configuration of the system may be non-deterministic. Thus, there can exist a relaxation of this traditional modal switching constraint, which in turn can simplify and accelerate planning \cite{hang2019pre}. We define a modal transition threshold vector, ${\rho}$, that defines an inflation of the constraint manifold along different axes, which inherently increases the potential volume of $\mathcal{M}_{\eta_1 \cap \eta_2}$ to develop a \textit{safe transfer region}, 
\begin{equation}\label{eq:modesRelaxed}
\mathcal{M}_{\eta \cap \eta'}^* = \{q \in \mathcal{Q} |-\rho<F^\eta(q) < \rho, -\rho<F^{\eta'}(q) < \rho \}
\end{equation}
\andy{The success of our finger gating platform in this work largely leverages this concept. Explicitly defining $\rho$, however, is difficult as it is system dependent, i.e. properties of the hand and the object determine how inflated the switching regions become. As a proof-of-concept analysis, we estimate $\rho$ via experimentation in Sec. \ref{sec:experiments} and will leave computational investigation for future work. } 
  \begin{figure*}[t]
      \centering
      \includegraphics[width = 0.98\textwidth]{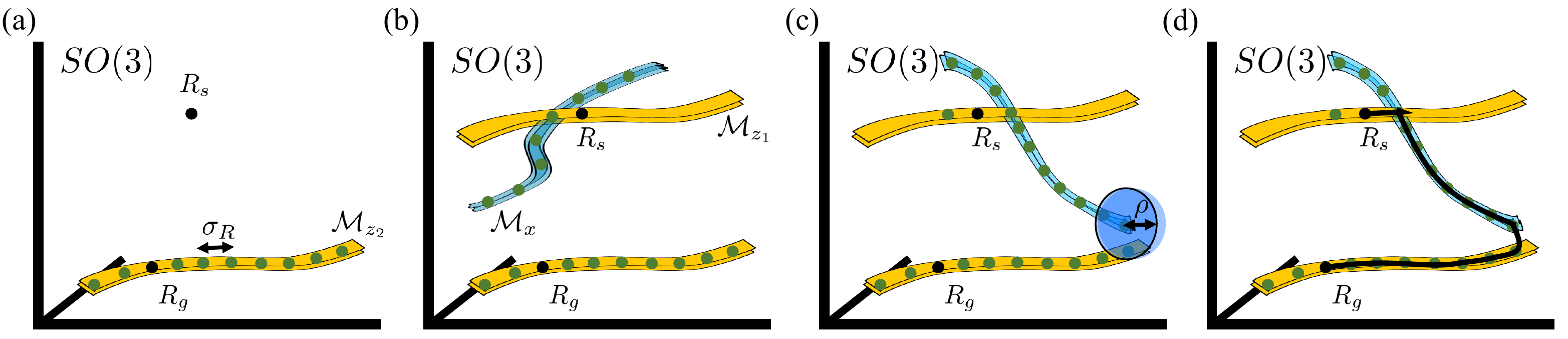}
       \vspace{-0.1in}
      \caption{(a) Our planning solution for $SO(3)$ begins by enumerating outward the goal orientation manifold by step size $\sigma_R$ and creating a KD-Tree. (b) After creation, a forward search outward is performed from the start orientation, $R_s$, combining a $z-$axis rotation first, then an $x-$axis rotation. (c) This search continues until a candidate orientation is within some distance $\rho$ from the expanded goal manifold. (d) Finally, the entire plan is enumerated from start to goal along the acquired trajectory with appropriate timestamps and step size $\sigma_R$.}
      \label{fig:so3Planning}
      \vspace{-.2in}
   \end{figure*}
\vspace{-.1in}
\section{ORIENTATION-BASED MOTION PLANNING}
\andy{We formally define the problem of orientation-based motion planning and the minimum number of actions required. Then, we utilize this as a guarantee for the completeness our planner's modal search transitions.}

\vspace{-.15in}
\subsection{Proper Rotations in $SO(3)$} \label{sec:proof}
\andy{\textit{Proper rotations} are sets of angles that can fully define any rotation in the group, $SO(3)$. Let's assume we have an object with some rotation, $R$. It is possible to transition $R$ to any rotation configuration via a combination of rotations along two orthogonal axes. Though there are six total combinations, we will instantiate our planner to focus on rotations along the $z-$axis ($R_z$) and $x-$axis ($R_x$) to serve as an example. }

\vspace{.03in}
\noindent \andy{\textbf{Theorem 1:} \label{theorem1}
\textit{Let the rotational configuration of the object be $R \in SO(3)$. Then there exists at least one $(\phi,\theta, \psi) \in [0,2\pi) \times [0,\pi] \times [0,2\pi)$ such that $R = R_z(\phi)R_x(\theta)R_z(\psi)$ \cite{craig2009introduction}.  }}

\vspace{-.14in}

\subsection{Planning Orientation Transitions} \label{Sec:alg}


\andy{Following this realization and the preliminaries in Sec. \ref{sec:prelims}, we develop a \textit{complete} multi-modal planning algorithm for finger gaiting that consists of three rotations along two orthogonal axes. 
If, for instance, all three orthogonal axes were available, the planning problem would be mostly trivial--simply transition the object along each axis as much as necessary to reach the goal. For the duration of this paper, we will refer to rotations about axes as being extrinsic, i.e. according to the hand frame of robot, and will adopt the notation that $R_x(\cdot)$ and $R_z(\cdot)$ are rotations of the object about the $x-$ and $z-$axes of the hand frame, or roll and yaw, respectively. }

\makeatletter
\def\BState{\State\hskip-\ALG@thistlm}
\makeatother
\setlength{\textfloatsep}{0pt}
\begin{algorithm}[thpb]
{\fontsize{9}{9} \selectfont
\caption{$SO3Plan(\cdot)$}\label{euclid}
\begin{algorithmic}[1]
	\renewcommand{\algorithmicrequire}{\textbf{Input:}}
	\renewcommand{\algorithmicensure}{\textbf{Output:}}
	\Require {$R_s, R_g, \rho, \sigma_R$}
	\Comment{object start orientation, object goal orientation, modal transition threshold, rotational step size}
	\Ensure {$\mathcal{P}$}
	\Comment{multi-modal plan}

    \State{$\Sigma = [0, \sigma_R, -\sigma_R, 2\sigma_R, -2\sigma_R, \cdots , \pi, -\pi] $}
    \Comment{candidate steps}
    \State{$\mathcal{M}_{z_2} \gets [R_z (\phi)R_g$ \textbf{ for }$\phi$ \textbf{ in }$\Sigma$]} 
    \Comment{expanded goal manifold}
    \State{$KD \gets KDTree.build(\mathcal{M}_{z_2})$}
    \Comment{KD-Tree of goal manifold}
    \For{$\psi$ \textbf{ in } $\Sigma$}
        \For{$\theta$ \textbf{ in } $\Sigma$}
            \State{$R^* = R_x(\theta)R_z(\psi)R_s$}
            \Comment{outward expanding search}
            \State{$\phi \gets KD.query\_closest\_config(R^*)$}
            \If{$ dist_R(R_z (\phi)R_g, R^*)< \rho$} \Comment{distance function}
                \State{\textbf{break loops}}
                \Comment{bidirectional trajectory connects}
            \EndIf
            \State {\textbf{end if}}
        \EndFor
        \State{\textbf{end for}}
    \EndFor
    \State{\textbf{end for}}
    \State{$\mathcal{P} \gets \textbf{planner}.enumerate(R_s, \sigma_R, \psi, \theta, \phi)$}   
    \NoNumber{\Comment{enumerate path outward according to connection (Fig. \ref{fig:so3Planning}(d))}}

    \State{\Return $\mathcal{P}$}
   
\end{algorithmic}
\label{Alg:rotation}
}
\end{algorithm}

\andy{Let's assume the capabilities of the robot enable the formulation of just two mode foliations, which are able to take an object in \textit{any} configuration and rotate about the $x-$ and $z-$axes, respectively. Given an object in start configuration $o_s = \{T_s \in \mathbb{R}^3, R_s \in SO(3)\}$, the objective of this planner is to reach a desired goal pose, $o_g= \{T_g \in \mathbb{R}^3, R_g \in SO(3)\}$. For now, we will disregard the translational positions, $T_s$ and $T_g$, and focus on rotation.}

\andy{As to accelerate computation, the proposed planning method is \textit{bi-directional} in that it builds a search trajectory outward from both, the start and goal configurations. More formally, there exists rotations such that $R_g = R_{z_2}(\phi)R_{x}(\theta)R_{z_1}(\psi)R_s$ where $R_{z_1}(\psi)$ is the first rotation about the $z-$axis, $R_{x}(\theta)$ is the rotation about the $x-$axis, and $R_{z_2}(\phi)$ is the second rotation about the $z-$axis. Thus, the goal of this planner is to solve for $\psi, \theta,$ and $\phi$ and enumerate a trajectory accordingly.}

\andy{Alg. \ref{Alg:rotation} begins with the backward pass by populating a $z-$axis rotation outward from $R_g$ with step size, $\sigma_R$, which is user-defined (Fig. \ref{fig:so3Planning}(a)). For our implementation, $\sigma_R$ can generally range anywhere from 0.5-5.0$\degree$ and can be adaptive according to system architectures. Let's say that these discretized rotations outward from $R_g$ can sufficiently define the entire manifold, $\mathcal{M}_{z_2}$, i.e. an entire rotation of the $z-$axis about $R_g$. From these configurations in $\mathcal{M}_{z_2}$, we can build a KD-Tree, $KD$, which is a data structure that enables efficient querying of the rotations included in this manifold.}

\andy{Now that the backwards direction is instantiated, the goal of the forward pass is to find rotations $R_x(\theta)R_{z_1}(\psi)R_s$ such that they can ``connect" to the manifold $\mathcal{M}_{z_2}$ represented by $KD$ within some distance threshold $\rho$ from \eqref{eq:modesRelaxed} (Fig. \ref{fig:so3Planning}(c)). As aforementioned, $\rho$ is the inflation factor for contact switching, and is generally system dependent. Here, we utilize a user-defined rotational distance function $dist_R(\cdot)$, where for our instantiation, we calculate the L$^2$-norm of the difference vector along the roll, pitch, and yaw dimensions of two rotations. Though, we note that there are other metrics available, e.g. \cite{huynh2009metrics3Drotations}. Tangibly, this forward pass sequentially searches through candidate values for $\psi$ and $\theta$ until the trajectory reaches the goal manifold $\mathcal{M}_{z_2}$, which is found by continually querying $KD$. Once this connection is found (Fig. \ref{fig:so3Planning}(d)), values for $\psi$, $\theta$, and $\phi$ are recorded, and a trajectory for plan $\mathcal{P}$ is enumerated with timestamps along step size $\sigma_R$ from start to goal. }

\vspace{-0.1in}
\section{Object Control}
\andy{Finger-gaiting is a dynamic and uncertain process; making and breaking contacts can often lead to slip which could further result in object ejection. To mitigate such occurrences, our solution is to continually replan and execute modal actions. More specifically, we rely on an online servoing-based framework so as to reach the desired goal. When the object enters a region of potential instability, such as being distant from the workspace center, a recovery phase comprised of translational modal actions is enacted to relocate the object. }

\subsubsection{Translational Planning}
\andy{Planning translations within-hand, presented in Alg. \ref{Alg:translation}, performs a simple, greedy visual servoing control approach. Generally, the algorithm calculates the translational difference, $\gamma$, along each of the coordinate axes from start position, $T_s$, and goal position, $T_g$. The planner then chooses the direction of largest difference and creates a plan to take a single step in that direction \cite{calli2017vision}.   }

\makeatletter
\def\BState{\State\hskip-\ALG@thistlm}
\makeatother
\begin{algorithm}[thpb]
{\fontsize{9}{9} \selectfont
\caption{$TranslationPlan(\cdot)$}\label{euclid}
\begin{algorithmic}[1]
	\renewcommand{\algorithmicrequire}{\textbf{Input:}}
	\renewcommand{\algorithmicensure}{\textbf{Output:}}
	\Require {$T_s, T_g, \sigma_T$}
	\Comment{object position, object goal position, step size}
	\Ensure {$\mathcal{P}$}
	\Comment{plan}
	
	\State {$\gamma.x \gets T_g.x -T_s.x$}
	\Comment{translational difference along $x-$axis}
	\State {$\gamma.y \gets T_g.y -T_s.y$}
	\Comment{translational difference along $y-$axis}
	\State {$\gamma.z \gets T_g.z -T_s.z$}
	\Comment{translational difference along $z-$axis}
	\State{$\mathcal{P} \gets \{\}$}
	\State{$\pi \gets max(|\gamma_x|,|\gamma_y|,|\gamma_z|)$}
	\Comment{direction of maximum deviation}
	\If {$\pi$ \textbf{is} $|\gamma_x|$}
	    \State{$\mathcal{P} \gets [ T_x+$sgn$(\gamma_x)\sigma_T, T_y, T_z]$}
    	\Comment{desired move $\pm T_x$}
	\ElsIf {$\pi$ \textbf{is} $|\gamma_y|$}
	    \State{$\mathcal{P} \gets [ T_x, T_y+$sgn$(\gamma_y)\sigma_T, T_z]$}
    	\Comment{desired move $\pm T_y$}
	\ElsIf {$\pi$ \textbf{is} $|\gamma_z|$}
	    \State{$\mathcal{P} \gets [ T_x, T_y, T_z+$sgn$(\gamma_z)\sigma_T]$}
    	\Comment{desired move $\pm T_z$}
   \EndIf
   \State{\Return $\mathcal{P}$}
\end{algorithmic}
\label{Alg:translation}
}
\end{algorithm}

\subsubsection{Orientation and Translation Control}
\andy{Our control approach combines both, Alg. \ref{Alg:rotation} and Alg. \ref{Alg:translation}. This method requires a goal threshold, $\tau$, a linear interpolant update parameter, $\lambda$, and \textit{a priori} knowledge of the hand's workspace center, $T_{center}$, for object recovery.}

\andy{In executing Alg. \ref{Alg:main}, orientation control is accomplished first. A plan $\mathcal{P}$ is found according to the object's current object rotation, $R$. The robot then takes a \textit{single} action along $\mathcal{P}$ via the \textbf{robot}$.mode\_action(\cdot)$ method. This system-specific function enacts a motor position step proportional to $\sigma_R$ along the desired mode. From this action, an object pose displacement, $\delta$, is calculated by taking the distance between the original orientation, $R$, and the new orientation, $R'$, after the action. This value $\delta$ serves as a differential update term along the orientation transition, which allows us to adaptively re-estimate $\sigma_R$ according to an interpolant update parameter $\lambda$. Since, as aforementioned, there is inherent uncertainty in the task of finger gaiting, this adaptive update to the transition step size $\sigma_R$ serves to aid in our plan's accuracy for the true motion of the hand-object system. The algorithm continues while $R$ is outside of the goal distance threshold $\tau_R$. Notably, a recovery phase (lines 12-16) is developed inside of this loop and is enacted when the object position, $T$, is distance $\tau_T$ outside of $T_{center}$. At this point, the system takes greedy transition steps towards the center of the workspace for recovery.  }

\andy{Alg. \ref{Alg:main} concludes by performing steps along $\mathcal{P}$ found from the $TranslationPlan(\cdot)$ method so as to reposition the object within-hand after object reorientation. We implement this method last in an attempt to decouple orientational and translational planning for the final pose. }

\makeatletter
\def\BState{\State\hskip-\ALG@thistlm}
\makeatother
\begin{algorithm}[thpb]
{\fontsize{9}{9} \selectfont
\caption{$ObjectControl(\cdot)$}\label{euclid}
\begin{algorithmic}[1]
	\renewcommand{\algorithmicrequire}{\textbf{Input:}}
	\renewcommand{\algorithmicensure}{\textbf{Output:}}
	\Require {$o_g$, $\rho$, $\sigma$, $\tau$, $\lambda$, $T_{center}$}

	\Comment{object goal pose, modal transition threshold vector, pose transition step size, goal pose reached threshold, interpolant step size, center of object workspace}
	\Ensure {$o_f$}
	\Comment{final object pose}
    
    \State {$T_g, R_g \gets o_g.T, o_g.R$}
    \Comment{translation and rotation of goal pose}
    \State {$o \gets \textbf{tracker}.perceive()$}
    \Comment{object pose $\in SE(3)$}
    \State {$T, R \gets o.T, o.R$}
    \Comment{translation and rotation of object pose}
    \NoNumber{}
    \While{$dist_R(R_g, R)>\tau_R$} \Comment{orientation control}

        \State{$\mathcal{P} \gets $\textbf{planner}$.SO3Plan(R, R_g, \rho, \sigma_R)$} 
        \Comment{Alg. \ref{Alg:rotation}}
        \State{\textbf{robot}$.mode\_action(\mathcal{P}, \sigma_R$)}
        \Comment{execute modal action}
        \State {$o' \gets \textbf{tracker}.perceive()$}
        \Comment{new object pose}
        \State {$T', R' \gets o'.T, o'.R$}
        \Comment{new object translation and rotation}
        \State{$\delta \gets dist_R(R',R)$}
        \Comment{size of last action step}
        \State{$\sigma_R \gets \sigma_R + \lambda(\sigma_R - \delta)$}
        \Comment{update rotational transition step}
        \State{$T, R \gets T', R'$}
        \While {$dist_T(T$,$T_{center}) > \tau_T$} \Comment{object recovery}
            \State{$\mathcal{P} \gets \textbf{planner}.TranslationPlan(T, T_{center}, \sigma_T)$}
            \State{\textbf{robot}$.mode\_action(\mathcal{P}, \sigma_T$)}
            \Comment{execute modal action}
            \State {$o \gets \textbf{tracker}.perceive()$}
            \Comment{object pose}
            \State {$T, R \gets o.T, o.R$}
            \Comment{object translation and rotation}
        \EndWhile
        \State{\textbf{end while}}
    \EndWhile
    \State{\textbf{end while}}
    \NoNumber{}
    \While {$dist_T(T_g, T) > \tau_T$} \Comment{translation planning}
        \State{$\mathcal{P} \gets \textbf{planner}.TranslationPlan(T, T_g, \sigma_T)$}
        \Comment{Alg. \ref{Alg:translation}}
        \State{\textbf{robot}$.mode\_action(\mathcal{P},\sigma_T$)}
        \Comment{execute modal action}
        \State {$o \gets \textbf{tracker}.perceive()$}
        \Comment{object pose}
        \State {$T, R \gets o.T, o.R$}
        \Comment{object translation and rotation}
    \EndWhile
    \State{$o_f \gets \textbf{tracker}.perceive()$}
    \Comment{final object pose}
    \State{\Return {$o_f$}}
\end{algorithmic}
\label{Alg:main}
}
\end{algorithm}




\subsubsection{Generalization and Practical Algorithm Modifications}
\andy{The authors present this solution as a generalized approach to planning for in-hand manipulation. Surely, the design and realization of the \textbf{robot}$.mode\_action(\cdot)$ method begs questions for system-specific scenarios, but can be found in a variety of ways: analytical modeling, dynamics learning, reinforcement learning, etc. While we do not focus on describing the underlying components of this function in this work, we largely leverage an energy-modeling technique outlined in \cite{morgan2020object} for determining predicted object transitions. }

\andy{Moreover, the proposed planner can be further accelerated by subtle variations to the aforementioned algorithms. First, $KD$ only needs to be computed once since the goal pose is static. Also, initialized step sizes for $\sigma$ can deviate between modes according to system-specific properties, e.g. when gravity affects the rotation for one mode more than another.  }


\vspace{-.1in}

\section{SYSTEM SETUP}

\begin{figure}
 \centering
 \includegraphics[width = 0.45\textwidth]{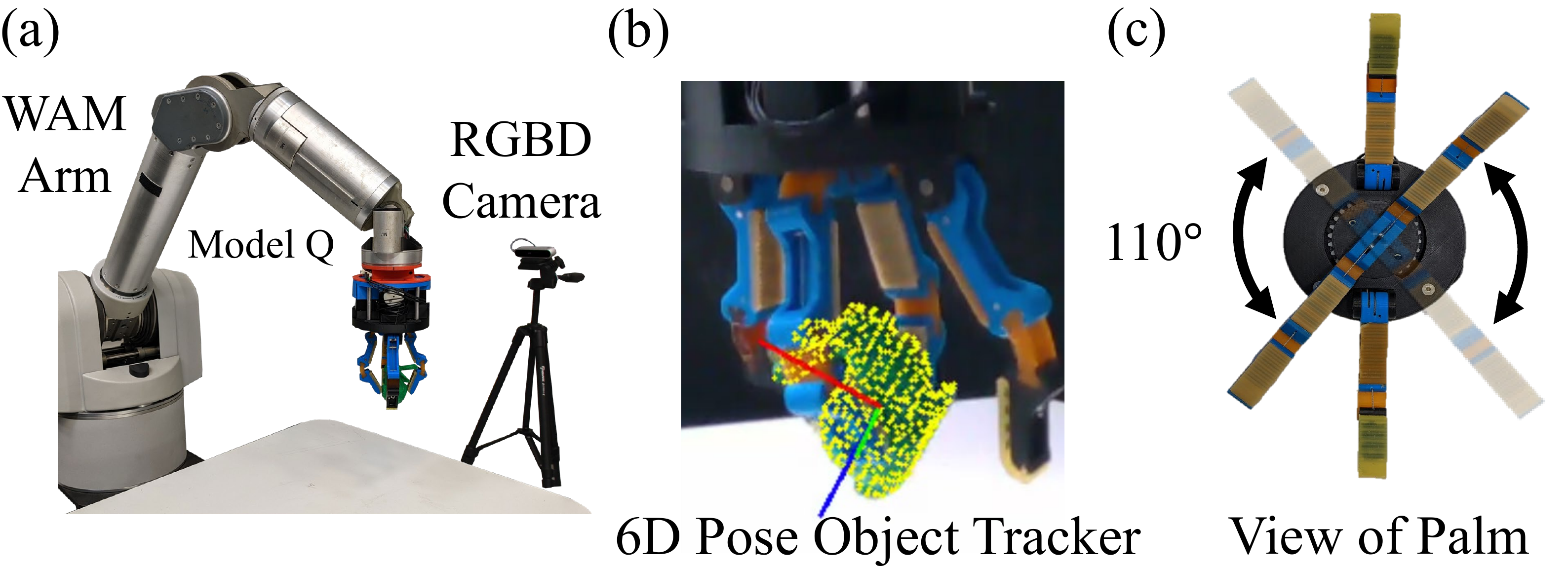}
  \vspace{-0.1in}
 \caption{(a) Our system is comprised of a RGBD camera fixated to the robot's environment. (b) During manipulation, the pose of the object, e.g. bunny, is tracked via a 6D pose object tracker. (c) A bottom view of the Yale Model Q illustrates the 110$\degree$ abduction capabilities of the differentially coupled fingers.}
 \label{fig:RobotSetup} 

 \end{figure}

\subsubsection{Robot Setup}

We utilize the open-source Model Q, an underactuated hand from the Yale OpenHand Project. The hand is equipped with four two-link fingers and four total actuators. A single motor controls the actuation of an opposing set of fingers with joints comprised of flexures, guided by a differential. This enables passive reconfigurability between these two fingers and into/out-of the plane of grasping. These two coupled fingers are connected to an actuated rotary joint located within the palm, allowing $110\degree$ rotation about the $z-$axis of the hand (Fig. \ref{fig:RobotSetup}). The final two motors serve to actuate the remaining two fingers individually via tendons. Being underactuated, the hand's joint configuration cannot be determined directly, as it is not equipped with joint encoders or tactile sensors. See \cite{ma2014underactuated} for additional design information. 


The hand is mounted on a 7-DOF Barrett WAM arm. 
\andy{External to the robot, an RGBD Intel Realsense D415 camera is calibrated to the robot's environment. Its imagery is processed through the low-latency (60Hz) 6D pose object tracker (Fig. \ref{fig:RobotSetup}). The tracker is robust to finger occlusion, has approximately $\pm$1.5$mm$/ 2$\degree$ of noise along any axis, and is trained solely with synthetic object data. See \cite{wen2020se} for additional information.}   

\begin{figure}[thpb]
  \centering
  \includegraphics[width = 0.44\textwidth]{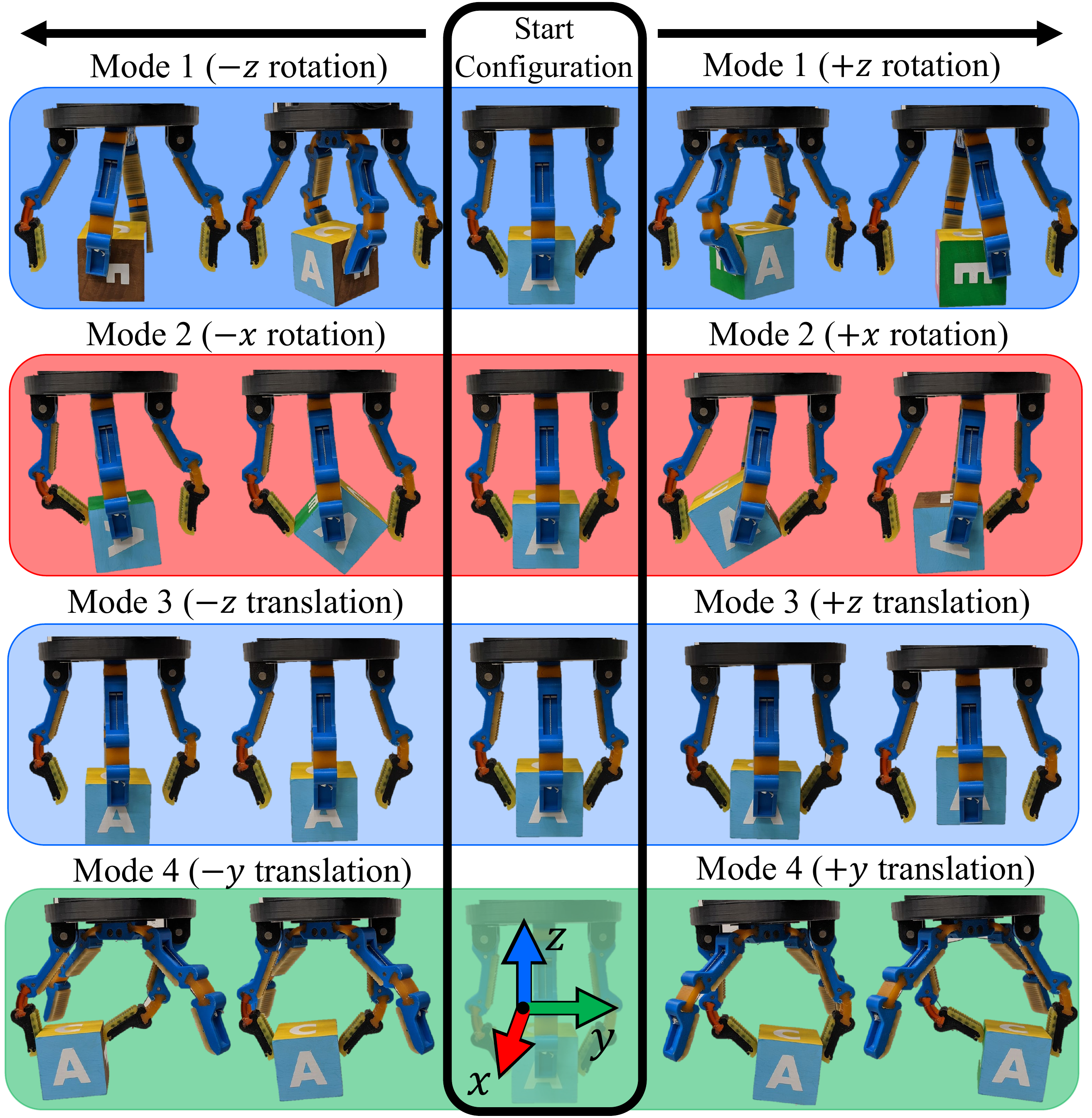}
   \vspace{-0.1in}
  \caption{The Yale OpenHand Model Q is capable of operating along four modes for fingertip-based manipulation. Modes are shown with transitions from the start configuration (center). }
  \label{fig:Modes}
   \vspace{-0.15in}
\end{figure}

\subsubsection{Mode Design} \label{sec:modeDesign}
As described in Sec. \ref{Sec:alg}, a series of rotations about two orthogonal axes allow an object to reach any orientation in $SO(3)$. Given the hand topology of the Model Q, an extrinsic $z-$axis rotation is easily possible via the palm's rotary joint. Moreover, an extrinsic $x-$axis rotation is also possible via coordinated finger motions; grasping with the differentially actuated opposing pair and pushing the object from either individually driven fingers as validated by \cite{morgan2020object}.

\andy{In addition to the two rotational motions, the hand topology furthermore allows for only two translational modes. The first can translate the object along the $z-$axis by squeezing the object towards the palm and regrasping, or by leveraging small amounts of coordinated slip, and simple $y-$axis translation is possible via a method described in \cite{calli2017vision} (Fig. \ref{fig:Modes}). Note that translation about the $x-$axis is not possible due to the actuation coupling of the rotary set of fingers.} 
\andy{By combining these two translational modes, we are able to provide a recovery phase for the object (Fig. 1, \ref{fig:BCTrans}), repositioning the object towards the middle of the workspace so that manipulation can safely continue. Note that $x-$axis translation is thus omitted.}

 \begin{figure}
 \centering
 \includegraphics[width = 0.38\textwidth]{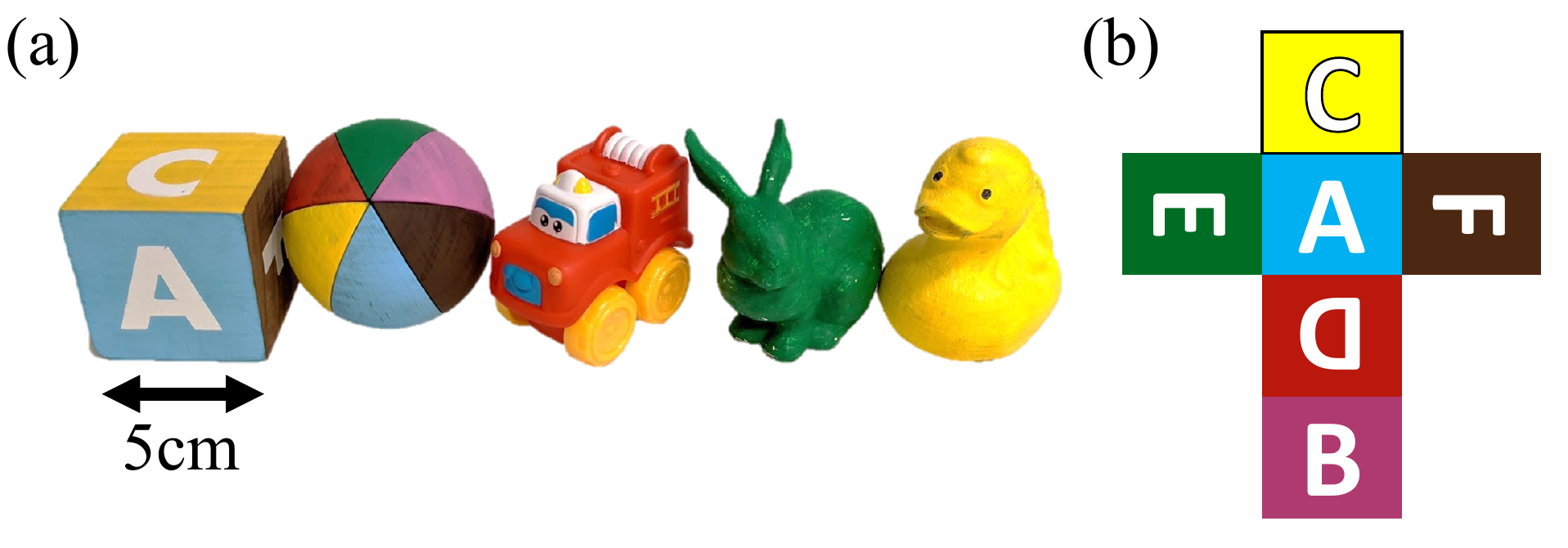}
  \vspace{-0.15in}
 \caption{(a) We experimentally validate our method with 5 different objects, namely a cube, sphere, toy truck, Stanford Bunny, and toy duck. (b) Tessellated faces of the cube show poses of the affixed letters. }
 \label{fig:objects picture} 
 \end{figure}

\vspace{-0.1in}
\section{EXPERIMENTS} \label{sec:experiments}

\begin{figure}
 \centering
 \includegraphics[width = 0.43\textwidth]{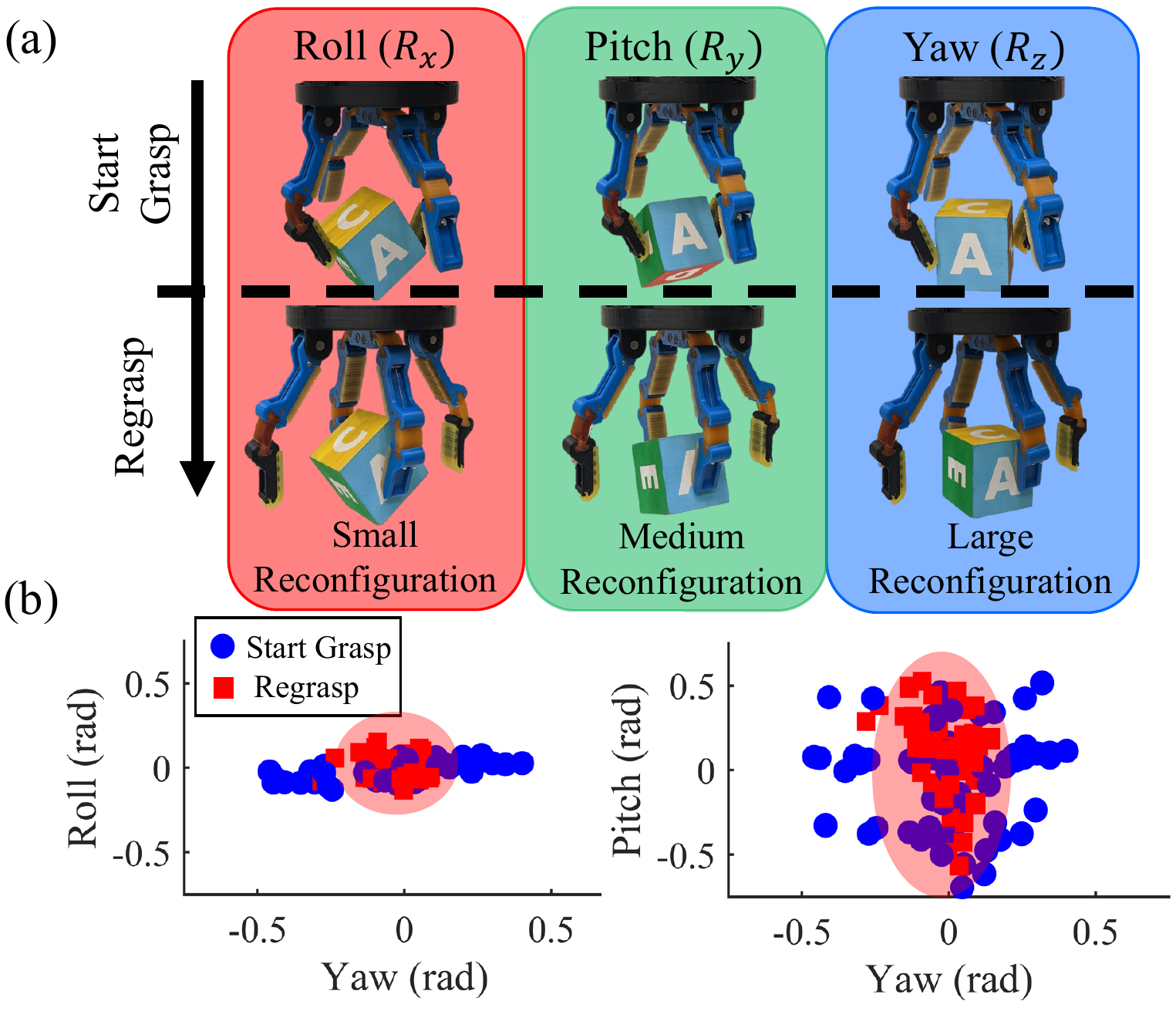}
  \vspace{-0.1in}
 \caption{The safety of mode transitions are attributed to the reconfigurability of the underactuated mechanism. (a) Starting rotations around different axes of the object elicit different amounts of reconfiguration upon regrasping. (b) This is particularly apparent in the yaw ($z-$axis) direction of the object, where the object follows the minimum energy configuration of the mechanism, and hence allowing us to estimate $\rho$ from Eq. \eqref{eq:modesRelaxed}.} 
 \label{fig:reconfiguration} 
 \end{figure}
 
\subsubsection{Safe Modes Characterization}
\andy{The passive adaptive nature of compliant hands is particularly beneficial for the task of finger gaiting. Simply, small errors associated with state estimation, modeling, and control can likely be dismissed by the system ``absorbing" the slack. It moreover determines the ``inflation" of switching regions, denoted by $\rho$. To help quantify and validate this adaptive nature, we evaluate how a \textit{regrasp} action, i.e. transferring the grasp between sets of opposing fingers, affects the resultant object configuration.}

\andy{Evaluated with the cube (Fig. \ref{fig:objects picture}), we perturb the object along each axis. Subsequently, the hand transfers the grasp to the opposite set of opposing fingers. In doing so, we record the amount of object reconfiguration along each axis via the object tracker. Notably, with the roll orientation of the object, we see a small amount of reconfiguration, but in the yaw orientation, we note the most reconfiguration (Fig. \ref{fig:reconfiguration}), due to the geometric properties of the cube. More specifically, from the start grasp (blue circle), to the final regrasped configuration (red square), the object tends to fall towards the minimum energy configuration of the system (red region) \cite{morgan2020object, Birchereabd2666}. 
We note that although there is a large variability in the object's start orientation ($\pm0.5$ radians), the hand is able to regrasp the object successfully in all cases. This underscores our \textit{safe modes} discussion, and generally allows us to estimate how large $\rho$ can be. For safety, we define $\rho=0.2$ radians. }
 \begin{figure*} [t]
 \centering
 \includegraphics[width = 0.9\textwidth]{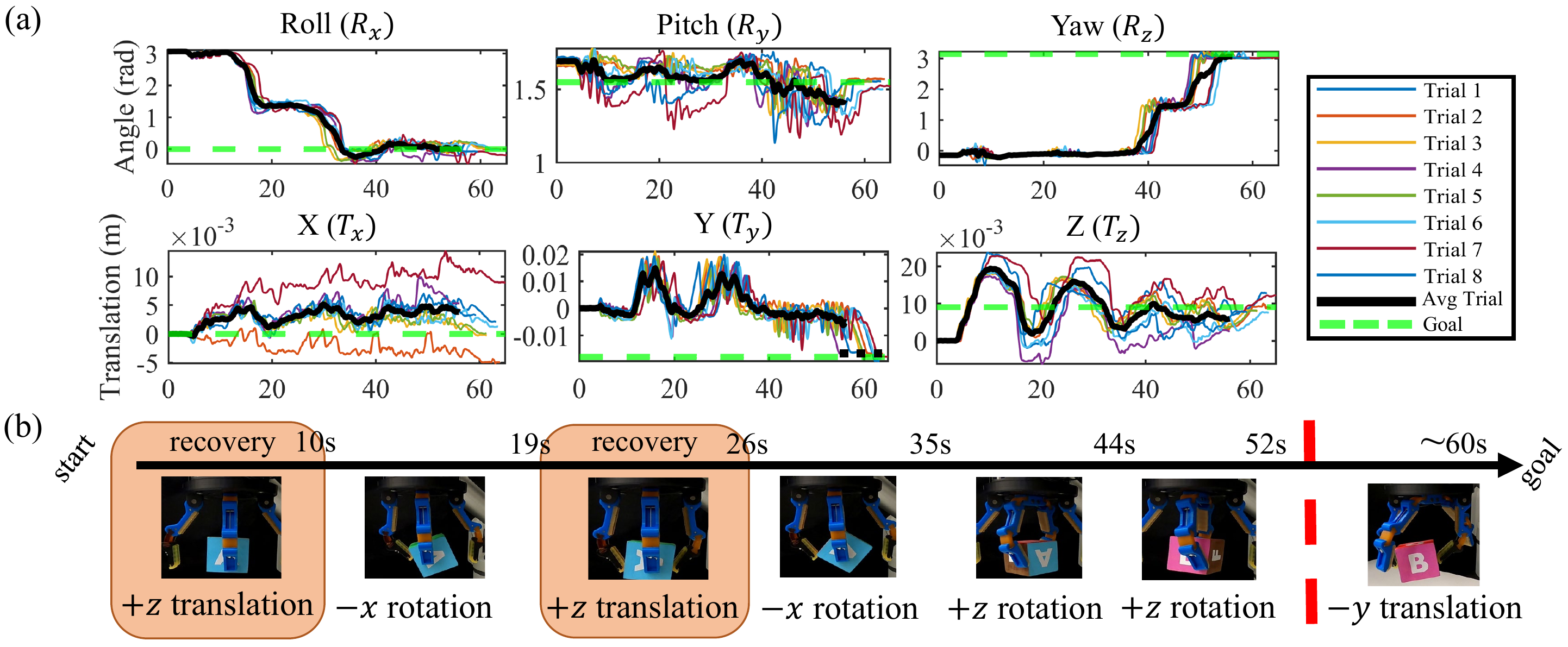}
 \vspace{-0.1in}
 \caption{We execute a single planned trajectory 8 times and record its repeatability. (a) Controlled dimensions of the object trajectory, such as roll and yaw, follow closely in all trials, where as uncontrolled dimensions, such as $x-$axis translation and pitch, are allowed to drift. (b) Modes are enacted at different times in the manipulation, including recovery phases when the object is not in the center of the graspable workspace.  }
 \label{fig:BCTrans} 
 \vspace{-.1in}
 \end{figure*}
 
  \begin{figure*} [t]
 \centering
 \includegraphics[width = 0.99\textwidth]{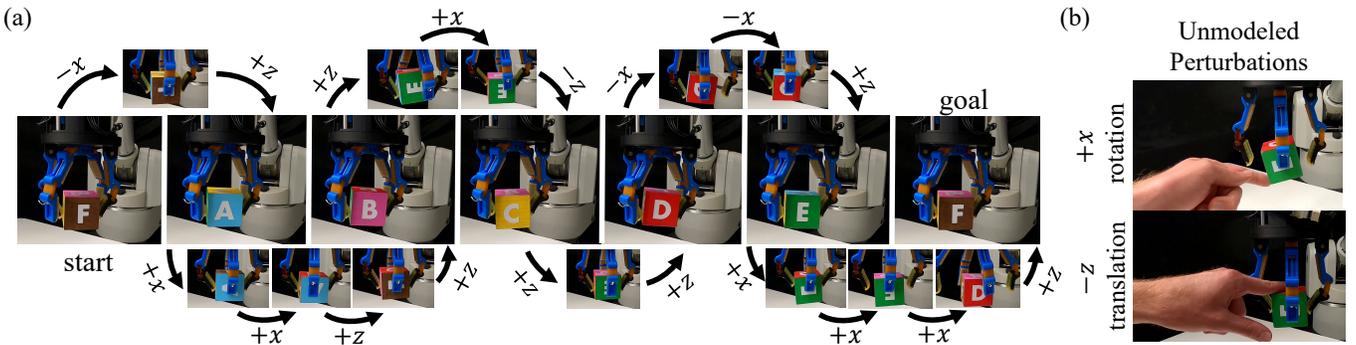}
  \vspace{-0.1in}
 \caption{We test system robustness by (a) following through an extended trajectory of cube faces F, A, B, C, D, E, and F once again, and (b) deliberately applying different perturbations to the object along controlled dimensions so as to require online replanning and recovery.}
 \label{fig:FABCDEF} 
 \vspace{-.2in}
 \end{figure*}

\subsubsection{Single Trajectory Execution}
We evaluate the repeatability of executing a single trajectory of the cube. Here, the goal of the task is to rotate the cube from face A to face B, and when completed, translate the object 1.8cm along the negative $y-$axis. The resultant executions, depicted in Fig. \ref{fig:BCTrans}(a), shows the average trajectory of the object over 8 different trials. During this task, we note how closely trajectories follow along the roll and yaw dimensions, as these are controlled via our safe modes. 
Object recovery occurs at similar times in the trials, where the system transitions the object along the positive $z-$axis before completing the $x-$axis rotations (Fig. \ref{fig:BCTrans}(b)). At the end of the orientation control sequence, the object is appropriately translated along the $y-$axis to reach its desired goal pose. The average trajectory data, depicted in solid black, does not extend past the 55 second mark due to one trial finishing earlier than others. The authors have, for clarity, extended the trajectory in the $y-$axis translation using a dotted line to illustrate the average final pose. 


\subsubsection{Continuous Goal Trajectories and Robustness}
We showcase the robustness of our method by evaluating both, an extended control trajectory and by disturbing the object pose during execution. As depicted in Fig. \ref{fig:FABCDEF}(a), we execute a trajectory starting on cube face F, transitioning through faces A,B,C,D,E and finally reaching the face F once again. In all cases, the object was able to reach the goal face within 8$\degree$.

Moreover, we illustrate our ability to replan control trajectories when unmodeled perturbations occur (Fig. \ref{fig:FABCDEF}(b)). 
\andy{At different points during manipulation, we randomly perturb the object pose 13 total times along the least constrained dimensions, the $x-$axis rotation and the $z-$axis translation.} Recovering and adapting to these occurrences over the course of a 205 second trajectory execution, the system completes the object reorientation task within 6$\degree$ of the goal face. 

 \begin{figure}
 \centering
 \includegraphics[width = 0.44\textwidth]{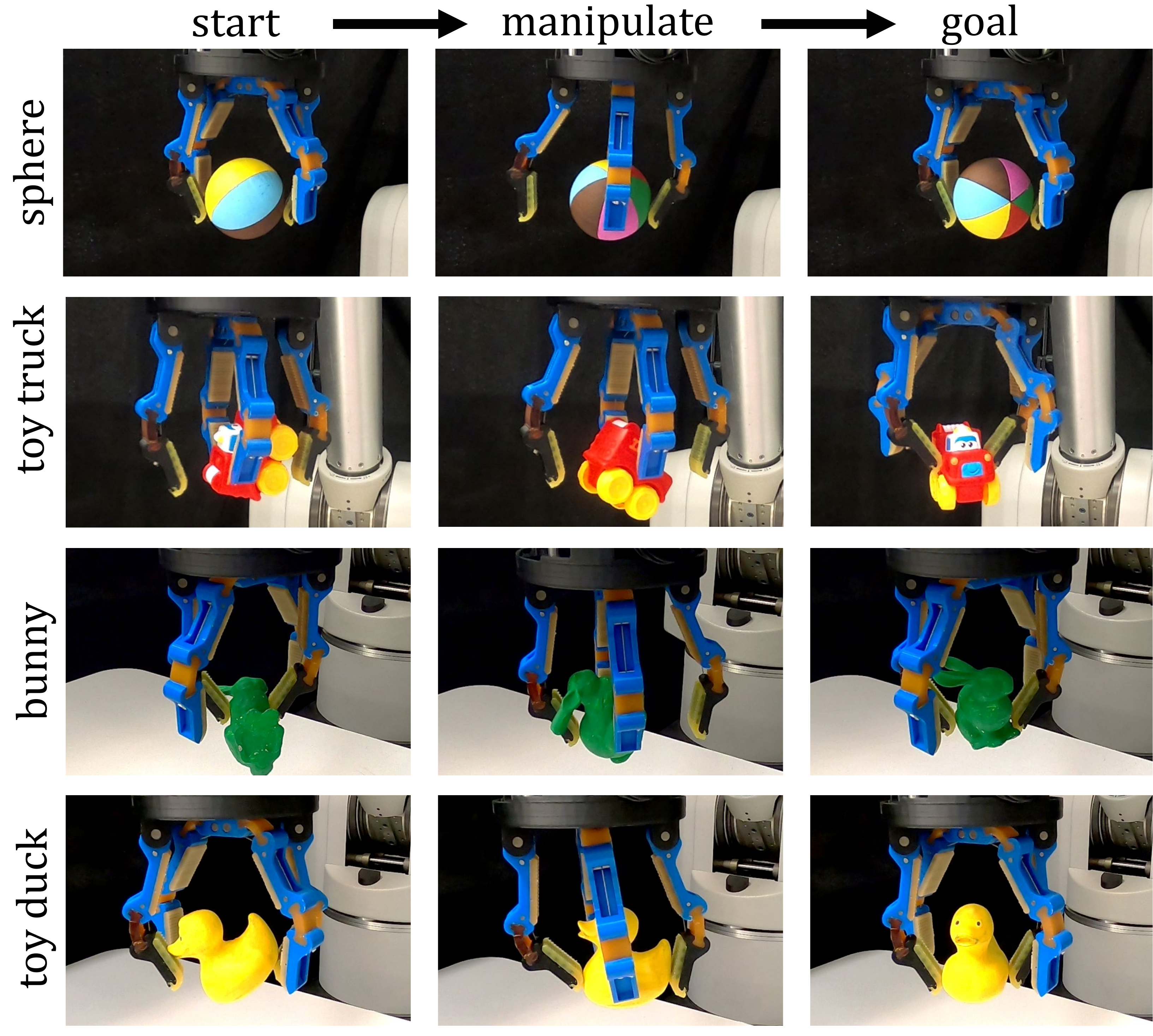}
  \vspace{-0.1in}
 \caption{Our manipulation planner is able to extend to objects of convex and non-convex geometries, and reach any orientation in $SO(3)$.}
 \label{fig:multipleObjects} 
 \end{figure}

\begin{figure}[thpb]
  \begin{center}
  \footnotesize{}
     \begin{tabular}{||c | c |c |c | c||} 
     \hline
     Obj.  &  $err$ (deg) & Plan time (s) & Total time (s) & Success \\ [0.5ex] 
     \hline\hline
     sphere & 4.6 & 13.2 & 153.1 & 5/5\\ 
     \hline
     truck & 4.2 & 10.8 & 112.3 & 5/5\\
     \hline
     bunny  & 9.5 & 14.0 & 128.0 & 2/5 \\
     \hline
     duck  & 8.9 & 7.1 & 73.2 & 3/5 \\
     \hline
    
\end{tabular}
\end{center}
    \vspace{-0.08in}
  \caption{Experimental evaluation of four different objects with error in final orientation, planning time, total time, and number of execution successes.}
  \label{tab:benchmarking}
\end{figure}

\subsubsection{Generalization to Object Geometry}
\andy{Finally, we test the transferability of our method by varying object geometries (Fig. \ref{fig:objects picture}). In doing so, recovery properties, such as $T_{center}$ from Alg. \ref{Alg:main}, were recalculated, and the 6D pose object tracker was retrained with new synthetic object data; other aforementioned parameters, including $\rho$, remained the same. }Depicted in Fig. \ref{fig:multipleObjects}, we evaluate simple, purely rotational trajectories about each of the objects. Notably, the sphere easily followed each of its guided reference trajectories, and was able to reach goal orientations within $\pm 5\degree$ along any axis. The toy truck, further illustrated our capabilities, transitioning to its goal configuration in approximately 110 seconds with 4 recovery phases. \andy{And finally, the Stanford Bunny and the toy duck were the most difficult to manipulate due to their non-convex properties. With the bunny, fingers would often get stuck behind its ears while making and breaking contact. Consequently, this occurrence caused the hand-object system to get stuck in specific object orientations and required restart. Moreover, during duck manipulation, the size of the duck's head affected the object's center of mass. This made some actions, such as $x-$axis rotation, difficult as the motion was then more dynamic in nature. Only a simple, two-rotation trajectory could be completed with the duck, resulting in a shorter execution time. The precision of these two manipulations was decreased to $\pm10\degree$ along any rotation axis and were consequently not quite as robust as the other objects. We provide benchmarking baselines for our experimentation in Fig. \ref{tab:benchmarking} according to guidelines outlined \cite{cruciani2020benchmarking}. Please refer to the supplementary video for experimental evaluations. }


\vspace{-0.1in}
\section{DISCUSSIONS AND FUTURE WORK}
In this work, we developed a robust and \textit{complete} solution to $SO(3)$ finger gait planning. While instantiated on an underactuated hand in this work, the methods described can generalize to other hand-object systems. We showcased our method with numerous tasks--highlighting the safety of our modes, the repeatability of our trajectories, the robustness of our planning and control approach, and our ability to generalize to new object scenarios. This work builds on a promising approach of vision-based control for compliant within-hand manipulation. The tasks completed in this letter extend beyond what has been possible in previous works. 

\andy{Although we showcase novel capabilities, there are some limitations. First, hand design plays a crucial role in our manipulation capabilities. We conceptualize an altered hand that would enable rotation about the $y-$axis, thus shortening trajectories. While our method may have relied less on data and/or advanced sensing, we note that our manipulation actions were significantly slower than some other related works \cite{andrychowicz2020learning, bhatt2021surprisingly}. Moreover, we discuss the ``inflation" parameter $\rho$ quite extensively, which begs for further theoretical investigation. To this end, the development of geometry-focused modal actions and transitions is an interesting avenue.
Lastly, the current tracking schema requires an object CAD model for training. We are interested in combining object-agnostic tracking methods \cite{wen2021bundletrack} together with the current object-agnostic controller for instant application to novel objects.}



\vspace{-0.1in}
\bibliographystyle{IEEEtran}
\bibliography{refs}


\end{document}